\documentclass[11pt,a4paper]{article}
\usepackage[hyperref]{naaclhlt2018}
\usepackage{times}
\usepackage{latexsym}
\usepackage{graphicx}
\usepackage{url}

\usepackage{float}
\usepackage{multirow}

\aclfinalcopy 

\title{A Report on the Complex Word Identification Shared Task 2018}

 \author{Seid Muhie Yimam$^1$, Chris Biemann$^1$, Shervin Malmasi$^2$, Gustavo H. Paetzold$^3$\\ 
 {\bf Lucia Specia}$^3$, {\bf Sanja \v{S}tajner}$^4$, {\bf Ana\"{i}s Tack}$^5$, {\bf Marcos Zampieri}$^6$\\
\vspace{-2mm}
\\ 
 $^1$University of Hamburg, Germany, $^2$Harvard Medical School, USA\\
 $^3$University of Sheffield, UK, $^4$University of Mannheim, Germany, \\
 $^5$Universit\'{e} catholique de Louvain and KU Leuven, Belgium\\
 $^6$University of Wolverhampton, UK\\
 \tt{yimam@informatik.uni-hamburg.de }\\
}

\date{}

\hypersetup{draft}

\begin{document}
\maketitle
\begin{abstract}
We report the findings of the second Complex Word Identification (CWI) shared task organized as part of the BEA workshop co-located with NAACL-HLT'2018. The second CWI shared task featured multilingual and multi-genre datasets divided into four tracks: English monolingual, German monolingual, Spanish monolingual, and a multilingual track with a French test set, and two tasks: binary classification and probabilistic classification. A total of 12 teams submitted their results in different task/track combinations and 11 of them wrote system description papers that are referred to in this report and appear in the BEA workshop proceedings. 
\end{abstract}

\section{Introduction}

The most common first step in lexical simplification pipelines is identifying which words are considered complex by a given target population \cite{shardlow2013comparison}. This task is known as complex word identification (CWI) and it has been attracting attention from the research community in the past few years.

In this paper we present the findings of the second Complex Word Identification (CWI) shared task organized as part of the thirteenth Workshop on Innovative Use of NLP for Building Educational Applications (BEA) co-located with NAACL-HLT'2018. The second CWI shared task follows a successful first edition featuring 21 teams organized at SemEval'2016 \cite{paetzold-specia:2016:SemEval1}. While the first CWI shared task targeted an English dataset, the second edition focused on multilingualism providing datasets containing four languages: English, German, French, and Spanish.

In an evaluation paper \cite{zampieri-EtAl:2017:NLPTEA}, it has been shown that the performance of an ensemble classifier built on top of the predictions of the participating systems in the 2016 task degraded, the more systems were added. The low performance of the CWI systems that competed in the first CWI task left much room for improvement and was one of the reasons that motivated us to organize this second edition. 


\subsection{Task Description}

The goal of the CWI shared task of 2018 is to predict which words challenge non-native speakers based on the annotations collected from both native and non-native speakers. To train their systems, participants received a labeled training set where words in context were annotated regarding their complexity. One month later, an unlabeled test set was provided and participating teams were required to upload their predictions for evaluation. More information about the data collection is presented in Section \ref{sec:datasets}.

Given the multilingual dataset provided, the CWI challenge was divided into four tracks:

\begin{itemize}
\item {\bf English monolingual CWI;}
\item {\bf German monolingual CWI;}
\item {\bf Spanish monolingual CWI; and}
\item {\bf Multilingual CWI with a French test set.}
\end{itemize}

\noindent For the first three tracks, participants were provided with training and testing data for the same language. For French, participants were provided only with a French test set and no French training data. 
In the CWI 2016, the task was cast as binary classification. To be able to capture complexity as a continuum, in our CWI 2018 shared task, we additionally included a probabilistic classification task. The two tasks are summarized as follows:

\begin{itemize}
\item {\bf Binary classification task:} Participants were asked to label the target words in context as complex (1) or simple (0).
\item {\bf Probabilistic classification task:} Participants were asked to assign the probability of target words in context being complex.
\end{itemize}

\noindent Participants were free to choose the task/track combinations they would like to participate in.

\section{Related Work}

Until the appearance of the CWI shared task of 2016, there was no manually annotated and verified CWI dataset. The 2016 shared task brought us one of the largest CWI datasets to that date, consisting of a total of 9,200 sentences manually annotated by 400 different non-native English speakers. In total, 200 sentences are used as a training set where each target is annotated by 20 annotators. The rest of the dataset (9,000 sentences) are used for test set where each target is annotated by a single annotator from the entire pool of 400 annotators.

The approaches used in the first SemEval 2016 Task 11: Complex Word Identification are described in Table \ref{tab:approaches}.

\begin{table*}[ht!]
\centering
\scalebox{0.82}{
  \begin{tabular}{lp{10cm}c}
  \hline
  \hline
  \bf Team & \bf Approach & \bf System Paper \\ \hline
  
  SV000gg & System voting with threshold and machine learning-based classifiers trained on morphological, lexical, and semantic features & \citep{paetzold-specia:2016:SemEval2} \\
  
  TALN & Random forests of lexical, morphological, semantic \& syntactic features & \citep{ronzano-EtAl:2016:SemEval} \\
  
  UWB & Maximum Entropy classifiers trained over word occurrence counts on Wikipedia documents & \citep{konkol:2016:SemEval} \\
  
  PLUJAGH & Threshold-based methods trained on Simple Wikipedia & \citep{wrobel:2016:SemEval} \\
  
  JUNLP & Random Forest and Naive Bayes classifiers trained over semantic, lexicon-based, morphological and syntactic features & \citep{mukherjee-EtAl:2016:SemEval} \\
  
  HMC & Decision trees trained over lexical, semantic, syntactic and psycholinguistic features & \citep{quijada-medero:2016:SemEval} \\
  
  MACSAAR & Random Forest and SVM classifiers trained over Zipfian features & \citep{zampieri-tan-vangenabith:2016:SemEval} \\
  
  Pomona & Threshold-based bagged classifiers with bootstrap re-sampling trained over word frequencies & \citep{kauchak:2016:SemEval} \\
  
  Melbourne & Weighted Random Forests trained on lexical/semantic features & \citep{brooke-uitdenbogerd-baldwin:2016:SemEval} \\
  
  IIIT & Nearest Centroid classifiers trained over semantic and morphological features & \citep{palakurthi-mamidi:2016:SemEval} \\
  
  LTG & Decision Trees trained over number of complex judgments & \citep{malmasi-dras-zampieri:2016:SemEval} \\
  
  MAZA & Ensemble methods various word frequency features & \citep{malmasi-zampieri:2016:SemEval} \\
  
  Sensible & Ensembled Recurrent Neural Networks trained over embeddings & \citep{nat:2016:SemEval} \\
  
  ClacEDLK & Random Forests trained over semantic, morphological, lexical and psycholinguistic features & \citep{davoodi-kosseim:2016:SemEval}\\
  
  Amrita-CEN & SVM classifiers trained over word embeddings and various semantic and morphological features & \citep{sp-kumar-kp:2016:SemEval} \\
  
  AI-KU & SVM classifier trained with word embeddings of the target and surrounding words & \citep{kuru:2016:SemEval} \\
  
  BHASHA & SVM and Decision Tree trained over lexical and morphological features & \citep{choubey-pateria:2016:SemEval} \\
  
  USAAR & Bayesian Ridge classifiers trained over a hand-crafted word sense entropy metric and language model perplexity & \citep{martinezmartinez-tan:2016:SemEval} \\
  
  CoastalCPH & Neural Network and Logistic Regression system trained over word frequencies and embedding & \citep{bingel-schluter-martinezalonso:2016:SemEval} \\
  \hline
  \end{tabular}
}
\caption{SemEval 2016 CWI -- Systems and approaches}
\label{tab:approaches}
\end{table*}


\section{Datasets}
\label{sec:datasets}

\begin{table}[ht!]
\centering
  \begin{tabular}{|l|c|c|c|}
   \hline
   \hline
   Language & Native & Non-native & Total \\
   \hline
   English & 134 & 49 & 183 \\
   German & 12 & 11 & 23 \\
   Spanish & 48 & 6 & 54 \\
   French & 10 & 12 & 22 \\
   \hline
 \end{tabular}
\caption{The number of annotators for different languages} 
\label{tab:numannotators}
\end{table}	

We have used the \emph{CWIG3G2} datasets from \cite{yimam-EtAl:2017:RANLP, yimam-EtAl:2017:I17-2} for the complex word identification (CWI) shared task 2018. The datasets are collected for multiple languages (\emph{English}, \emph{German}, \emph{Spanish}). The English datasets cover different text genres, namely \emph{News} (professionally written news), \emph{WikiNews} (news written by amateurs), and \emph{Wikipedia articles}. Below, we will briefly describe the annotation process and the statistics of collected datasets. For detail explanation of the datasets, please refer to the works of \citet{yimam-EtAl:2017:RANLP, yimam-EtAl:2017:I17-2}

Furthermore, to bolster the cross-lingual CWI experiment, we have collected a CWI dataset for French. The French dataset was collected through the same method used for the CWIG3G2 corpus \citep{yimam-EtAl:2017:RANLP, yimam-EtAl:2017:I17-2}. The dataset contains Wikipedia texts extracted from a comparable simplified corpus collected by \citet{brouwersSyntactic2014}. Similar to CWIG3G2, for each article, all paragraphs containing between 5 and 10 sentences were extracted. From this pool of paragraphs, only the best paragraph was selected via a ranking procedure maximizing sentence length and lexical richness, and minimizing the ratio of named entities and foreign words. From this large selection of best paragraphs per article, an optimal subset of 100 paragraphs was then selected using a greedy search procedure similar to that of \citet{tackEvaluating2016}, minimizing the vocabulary overlap between pairs of paragraphs using the Jaccard coefficient. Finally, a random test split of 24 paragraphs was selected to be annotated.

\subsection{Annotation Process}
Annotations were collected using the Amazon Mechanical Turk (MTurk). Instead of showing a single sentence, we presented 5 to 10 sentences to the annotator in a single HIT (Human Intelligence Task) and requested them to highlight words or phrases that could pose difficulty in understanding the paragraph. The annotation system is unique in many aspects such as: 1) The instruction makes clear that the annotators should assume a given target reader such as children, language learners or people with reading impairments. 2) A bonus reward is offered when the user's selection matches at least half of the other annotations to encourage extra care during the complex word or phrase (CP) selection. 3) The maximum number of annotations allowed is limited to 10 so that we could prohibit an arbitrarily large number of selections intending to attain the bonus reward. 4) For the English dataset, more than 20 annotators were able to annotate the same HIT, among which are at least 10 native English speakers and 10 non-native English speakers so that it is possible to investigate if native and non-native speakers have different CWI needs. 5) Complex words are not pre-highlighted, as in previous contributions, so that annotators are not biased to the pre-selection of the complex phrases. 6) In addition to single words, we allowed the annotation of multi-word expressions (MWE), up to a size of 50 characters.

Table \ref{tab:numannotators} shows the total, native, and non-native number of annotators that participated in the annotation task.
\begin{table}[ht!]
\centering
  \begin{tabular}{|l|c|c|c|}
   \hline
   \hline
   Language & Train & Dev & Test \\
   \hline
   English & 27,299 & 3,328 & 4,252 \\
   German & 6,151 & 795 & 959 \\
   Spanish & 13,750 & 1,622 & 2,233 \\
   French & - & - & 2,251 \\
   \hline
 \end{tabular}
\caption{The number of instances for each training, development and test set } 
\label{tab:anostats}
\end{table}

\subsection{Analysis of Collected Datasets}

Table \ref{tab:anostats} shows statistics of the datasets for the English (combinations of three genres), German, Spanish and French (test set only) CWI tasks.

\noindent An analysis of the English dataset shows that around 90\% of complex phrases have been selected by at least two annotators (both native and non-native). When separated by language, the percentage of agreements decreases to 83\% at the lowest. This might be because native and non-native annotators have a different perspective what is a complex phrase. Furthermore, we have seen that native annotators agree more within their group (84\% and above) than non-native speakers (83\% and above). We also see that the absolute agreement between native and non-native annotators is very low (70\%), which further indicates that the two user groups might have different CWI needs.

For the German annotation task, we have fewer annotators than the other languages. As it can be seen from Table \ref{tab:numannotators}, there are more native annotators, but they participate on fewer HITs than the non-native annotators (on average, 6.1 non-native speakers and 3.9 native speakers participated in a HIT). Unlike the English annotation task, non-native annotators have a higher inter-annotator agreement (70.66\%) than the native annotators (58.5\%). 

The Spanish annotation task is different from both the English and the German annotation tasks since its annotations come almost exclusively from native annotators. In general, Spanish annotators have shown lower agreements than the English and German annotators. Also the Spanish annotators highlight more MWEs than the English and German annotators.

Regarding the French annotation task, we observe a comparable distribution in the number of native and non-native annotators compared to the German annotation task (Table~\ref{tab:numannotators}). There were slightly more non-native participants than native ones, but the number of native annotators who completed the same number of HITs was considerably larger. This means that although there were more non-native participants, they did not participate equally in all HITs.

\begin{table}[ht!]
\centering
\begin{tabular}{|l|rr|rr|rr|}
\hline
\hline
{}   & \multicolumn{2}{c|}{Train} & \multicolumn{2}{c|}{Dev} & \multicolumn{2}{c|}{Test} \\
\hline
{}   & \#   & \%        & \# & \%          & \# & \% \\
\hline
EN & 11,253 & 41        & 1,388 & 42        & 1,787 & 42 \\
DE & 2,562 & 42        & 334 & 42         & 376 & 39\\
ES & 5,455 & 40        & 653 & 40         & 907 & 41\\
FR & \multicolumn{2}{c|}{-}   & \multicolumn{2}{c|}{-}  & 657 & 29 \\
\hline
\end{tabular}
\caption{\label{tab:numcomplex}The number (\#) and ratio (\%) of complex instances per language} 
\end{table}

A striking difference that can be observed in the French dataset pertains to the proportion of identified complex words. Compared to the other languages, we have a considerably lower relative count of complex instances (Table~\ref{tab:numcomplex}). However, this does not necessarily mean that the texts were simpler for French than for the other languages. Looking at the proportion of MWEs annotated as complex (Table~\ref{tab:numngrams}), we observe that the French dataset contains more MWE annotations than single words compared to the other datasets. One plausible explanation for this could be attributed to the limitation of allowing at most 10 unique annotations per HIT in MTurk. Indeed, a number of annotators highlighted the fact that they sometimes found more than 10 possible annotations of complex words. As a result, in order to account for all of these possibilities, the annotators sometimes grouped nearly adjacent single complex words as one sequence, leading to a larger relative proportion of MWE (3-gram+) annotations. 
Another explanation for this disparity could be attributed to the lower number of annotators for French compared to English or Spanish. If we had had a similar number of annotators for French, we would probably also have obtained a more varied sample and hence a higher relative amount of different complex word annotations.

\begin{table}[ht!]
\centering
\resizebox{\linewidth}{!}{%
\begin{tabular}{|ll|rrr|r|}
\hline
\hline
{} & {} & 1-gram & 2-gram & 3-gram+ & total \\
\hline
\multirow{2}{*}{EN} & \# & 10,676 & 2,760 & 992   & 14,428 \\
{}         & \% & 74.00 & 19.13 & 6.87  & \\
\hline
\multirow{2}{*}{DE} & \# & 2,770 & 307  & 195   & 3,272 \\
{}         & \% & 84.66 & 9.38  & 5.96  & \\
\hline
\multirow{2}{*}{ES} & \# & 4,712 & 1,276 & 1,027  & 7,015 \\
{}         & \% & 67.17 & 18.19 & 14.64  & \\
\hline
\multirow{2}{*}{FR} & \# & 414  & 118  & 125   & 657 \\
{}         & \% & 63.01 & 17.96 & 19.03  &   \\
\hline
\end{tabular}
}
\caption{\label{tab:numngrams}The distribution of single and MWE annotations of complex words per language} 
\end{table}

\section{System Descriptions and Results}

\begin{table*}[ht!]
\centering
\scalebox{.88}{
   \begin{tabular}{|l|c|c|l|c|c|l|c|c|}
   \hline
   \hline
		{\bf News} & F-1 & Rank & {\bf WikiNews} & F-1 & Rank & {\bf Wikipedia} & F-1 & Rank\\
		\hline
        
		Camb & 0.8736 & 1 & Camb & 0.84 & 1 & Camb & 0.8115 & 1\\
		Camb & 0.8714 & 2 & Camb & 0.8378 & 2 & NILC & 0.7965 & 2\\
		Camb & 0.8661 & 3 & Camb & 0.8364 & 4 & UnibucKernel & 0.7919 & 3\\
		ITEC & 0.8643 & 4 & Camb & 0.8378 & 3 & NILC & 0.7918 & 4\\
		ITEC & 0.8643 & 4 & NLP-CIC & 0.8308 & 5 & Camb & 0.7869 & 5\\
		TMU & 0.8632 & 6 & NLP-CIC & 0.8279 & 6 & Camb & 0.7862 & 6\\
		ITEC & 0.8631 & 7 & NILC & 0.8277 & 7 & SB@GU & 0.7832 & 7\\
		NILC & 0.8636 & 5 & NILC & 0.8270 & 8 & ITEC & 0.7815 & 8\\
		NILC & 0.8606 & 9 & NLP-CIC & 0.8236 & 9 & SB@GU & 0.7812 & 9\\
		Camb & 0.8622 & 8 & CFILT\_IITB & 0.8161 & 10 & UnibucKernel & 0.7804 & 10\\
		NLP-CIC & 0.8551 & 10 & CFILT\_IITB & 0.8161 & 10 & Camb & 0.7799 & 11\\
		NLP-CIC & 0.8503 & 12 & CFILT\_IITB & 0.8152 & 11 & CFILT\_IITB & 0.7757 & 12\\
		NLP-CIC & 0.8508 & 11 & CFILT\_IITB & 0.8131 & 12 & CFILT\_IITB & 0.7756 & 13\\
		NILC & 0.8467 & 15 & UnibucKernel & 0.8127 & 13 & CFILT\_IITB & 0.7747 & 14\\
		CFILT\_IITB & 0.8478 & 13 & ITEC & 0.8110 & 14 & NLP-CIC & 0.7722 & 16\\
		CFILT\_IITB & 0.8478 & 13 & SB@GU & 0.8031 & 15 & NLP-CIC & 0.7721 & 17\\
		CFILT\_IITB & 0.8467 & 14 & NILC & 0.7961 & 17 & NLP-CIC & 0.7723 & 15\\
		SB@GU & 0.8325 & 17 & NILC & 0.7977 & 16 & NLP-CIC & 0.7723 & 15\\
		SB@GU & 0.8329 & 16 & CFILT\_IITB & 0.7855 & 20 & SB@GU & 0.7634 & 18\\
		Gillin Inc. & 0.8243 & 19 & TMU & 0.7873 & 19 & TMU & 0.7619 & 19\\
		Gillin Inc. & 0.8209 & 24 & SB@GU & 0.7878 & 18 & NILC & 0.7528 & 20\\
		Gillin Inc. & 0.8229 & 20 & UnibucKernel & 0.7638 & 23 & UnibucKernel & 0.7422 & 24\\
		Gillin Inc. & 0.8221 & 21 & hu-berlin & 0.7656 & 22 & hu-berlin & 0.7445 & 22\\
		hu-berlin & 0.8263 & 18 & SB@GU & 0.7691 & 21 & SB@GU & 0.7454 & 21\\
		Gillin Inc. & 0.8216 & 22 & LaSTUS/TALN & 0.7491 & 25 & UnibucKernel & 0.7435 & 23\\
		UnibucKernel & 0.8178 & 26 & LaSTUS/TALN & 0.7491 & 25 & LaSTUS/TALN & 0.7402 & 25\\
		UnibucKernel & 0.8178 & 26 & SB@GU & 0.7569 & 24 & LaSTUS/TALN & 0.7402 & 25\\
		CFILT\_IITB & 0.8210 & 23 & hu-berlin & 0.7471 & 26 & NILC & 0.7360 & 26\\
		CFILT\_IITB & 0.8210 & 23 & Gillin Inc. & 0.7319 & 28 & hu-berlin & 0.7298 & 27\\
		hu-berlin & 0.8188 & 25 & Gillin Inc. & 0.7275 & 30 & CoastalCPH & 0.7206 & 28\\
		UnibucKernel & 0.8111 & 28 & Gillin Inc. & 0.7292 & 29 & LaSTUS/TALN & 0.6964 & 29\\
		NILC & 0.8173 & 27 & Gillin Inc. & 0.7180 & 31 & Gillin Inc. & 0.6604 & 30\\
		LaSTUS/TALN/TALN & 0.8103 & 29 & LaSTUS/TALN & 0.7339 & 27 & Gillin Inc. & 0.6580 & 31\\
		LaSTUS/TALN & 0.8103 & 29 & Gillin Inc. & 0.7083 & 32 & Gillin Inc. & 0.6520 & 32\\
		LaSTUS/TALN & 0.7892 & 31 & UnibucKernel & 0.6788 & 33 & Gillin Inc. & 0.6329 & 33\\
		UnibucKernel & 0.7728 & 33 & SB@GU & 0.5374 & 34 & SB@GU & 0.5699 & 34\\
		SB@GU & 0.7925 & 30 & - & - & - & CoastalCPH & 0.5020 & 35\\
		SB@GU & 0.7842 & 32 & - & - & - & LaSTUS/TALN & 0.3324 & 36\\
		LaSTUS/TALN & 0.7669 & 34 & - & - & - & - & - & -\\
		UnibucKernel & 0.5158 & 36 & - & - & - & - & - & -\\
		SB@GU & 0.5556 & 35 & - & - & - & - & - & -\\
		LaSTUS/TALN & 0.2912 & 37 & - & - & - & - & - & -\\
		LaSTUS/TALN & 0.1812 & 38 & - & - & - & - & - & -\\
		LaSTUS/TALN & 0.1761 & 39 & - & - & - & - & - & -\\
        \hline
		Baseline & 0.7579 & - & Baseline & 0.7106 & - & Baseline & 0.7179 & -\\
		    \hline
   \end{tabular}
   }
 \caption{Binary classification results for the monolingual English tracks.} 
\label{tab:binresultsmono}
\end{table*}
	
\begin{table*}[ht!]
\centering
\scalebox{0.90}{
   \begin{tabular}{|l|c|c|l|c|c|l|c|c|}
   \hline
   \hline
		{\bf German} & F-1 & Rank & {\bf Spanish}  & F-1 & Rank & {\bf French} & F-1 & Rank \\
		\hline
		TMU & 0.7451 & 1 & TMU & 0.7699 & 1 & CoastalCPH & 0.7595 & 1\\
		SB@GU & 0.7427 & 2 & ITEC & 0.7637 & 3 & TMU & 0.7465 & 2\\
		hu-berlin & 0.6929 & 4 & NLP-CIC & 0.7672 & 2 & SB@GU & 0.6266 & 3\\
		SB@GU & 0.6992 & 3 & CoastalCPH & 0.7458 & 5 & SB@GU & 0.6130 & 4\\
		CoastalCPH & 0.6619 & 5 & CoastalCPH & 0.7458 & 5 & hu-berlin & 0.5738 & 6\\
		Gillin Inc. & 0.5548 & 10 & NLP-CIC & 0.7468 & 4 & SB@GU & 0.5891 & 5\\
		Gillin Inc. & 0.5459 & 11 & NLP-CIC & 0.7419 & 6 & hu-berlin & 0.5343 & 7\\
		Gillin Inc. & 0.5398 & 12 & SB@GU & 0.7281 & 7 & hu-berlin & 0.5238 & 8\\
		Gillin Inc. & 0.5271 & 14 & SB@GU & 0.7259 & 8 & hu-berlin & 0.5124 & 9\\
		Gillin Inc. & 0.5275 & 13 & CoastalCPH & 0.7238 & 9 & - & - & - \\
		CoastalCPH & 0.6078 & 6 & hu-berlin & 0.7080 & 11 & - & - & - \\
		CoastalCPH & 0.5818 & 7 & CoastalCPH & 0.7153 & 10 & - & - & - \\
		CoastalCPH & 0.5778 & 8 & Gillin Inc. & 0.6804 & 13 & - & - & - \\
		CoastalCPH & 0.5771 & 9 & Gillin Inc. & 0.6784 & 14 & - & - & - \\
		- & - & - & Gillin Inc. & 0.6722 & 15 & - & - & - \\
		- & - & - & Gillin Inc. & 0.6669 & 16 & - & - & - \\
		- & - & - & Gillin Inc. & 0.6547 & 17 & - & - & - \\
		- & - & - & CoastalCPH & 0.6918 & 12 & - & - & - \\
        \hline
		Baseline & 0.7546 & - & Baseline & 0.7237 & - & Baseline & 0.6344 & - \\
    \hline
   \end{tabular}
   }
 \caption{Binary classification results for the multilingual German, Spanish and French tracks.} 
\label{tab:binresultsmulti}
\end{table*}

In this section, we briefly describe the systems from all 11 teams that have participated in the 2018 CWI shared task and wrote a system description paper to be presented at the BEA conference. Table \ref{tab:binresultsmono} and \ref{tab:binresultsmulti} shows the results of all systems for the monolingual and multilingual binary classification tasks while Table \ref{tab:Probresultsmono} and \ref{tab:Probresultsmult} presents the probabilistic classification results for the monolingual and multilingual tracks.

\subsection{Baseline Systems}
For both the binary and probabilistic classification tasks, we build a simple baseline system that uses only the most basic features described in \citet{yimam-EtAl:2017:RANLP, yimam-EtAl:2017:I17-2}, namely only frequency and length features. The Nearest Centroid classifier and the Linear Regression algorithms from the scikit-learn machine learning library are used for the binary and probabilistic classification tasks resp. For the binary classification task, we have used the accuracy and macro-averaged F1 evaluation metrics. For the probabilistic classification task, the Mean Absolute Error (MAE) measure is used. The baseline results are shown in Table \ref{tab:binresultsmono}, \ref{tab:binresultsmulti}, \ref{tab:Probresultsmono}, and  \ref{tab:Probresultsmult} for the monolingual and multilingual tracks.

\subsection{Shared Task Systems}

\textbf{UnibucKernel} The UnibucKernel \cite{syspaper1} team participated on the monolingual CWI shared task, specifically on the \textsc{News}, \textsc{WikiNews}, and \textsc{Wikipedia} domain datasets. The pipeline consists of feature extraction, computing a kernel matrix and applying an SVM classifier.

The feature sets include low-level features such as character n-grams, and high-level features such semantic properties extracted from lexical resources and word embeddings. The low-level features were extracted based on the target complex word, and include count of characters, count of vowels, count of consonants, count of repeating characters, and count of character n-grams (up to 4 characters).

The first set of word embedding features take into account the word's context which is obtained by computing the cosine similarity between the complex word and each of the other words in the sentence (minimum, maximum and mean similarity values are used). Furthermore, sense embeddings are used, which are computed based on WordNet synsets. Lastly, using word embeddings, additional features were designed based on the location of the complex word in a dimensionally reduced embedding space. For this, they used PCA to reduce the dimension of the embeddings from 300 to 2 dimensions. 

Once features are extracted, kernel-based learning algorithms are employed. For the binary classification setup, the SVM classifiers based on the Lib-SVM were used. For the regression setup, they used v-Support Vector Regression (v-SVR). For both setups, different parameters were tuned using the development dataset. 

\textbf{SB@GU} systems \cite{syspaper2} are adapted from a previous system, which was used to classify Swedish words into different language proficiency levels and participated on the multilingual binary classification part of the shared task. For each target word or MWE, the following set of feature categories were extracted: 1) count and word form features such as length of the target, number of syllables, n-gram probabilities based on Wikipedia, binary features such as ``is MWE'' or ``is number'', and so on 2) morphological features, mainly part-of-speech tag and suffix length, 3) semantic features, such as the number of synsets, number of hypernyms, and number of hyponyms, 4) context features, like topic distributions and word embeddings, and 5) psycholinguistic features, such as British National Corpus frequency, reaction time, bigram frequency, trigram frequency, and so on. For MWE, they averaged the feature values for each word in them.

For English datasets, experiments are conducted with context-free, context-only and context-sensitive features, mainly by excluding word embeddings, using only word embeddings, and combining all features explained above respectively. Classifiers such as Random Forest, Extra Trees, convolutional networks, and recurrent convolutional neural networks were tested. Furthermore, feature selection is performed using the SelectFromModel feature selection method from scikit-learn library. The best performing features includes word frequency, word sense and topics, and language model probabilities. 

For the German, Spanish, and French datasets, features such as character-level n-grams were extracted from n-gram models trained on Wikipedia. For the French dataset, the n-gram models from English, German and Spanish were used to obtain n-gram probabilities of each entry. They configured two setups to extract features for the French dataset: 1) Uses English, German and Spanish classifiers and apply majority voting to get the final label, 2) Uses only the Spanish classifier as French and Spanish are both Romance languages. 

An Extra Tree classifier with 1000 and 500 estimators was their best classifier. 

\textbf{hu-berlin} The systems \cite{syspaper3} mainly explored the use of character n-gram features using a multinomial Naive Bayes classifier specifically designed for the multilingual binary classification task. For each target word, all the character n-grams of a given length and their frequencies were extracted and the target word was represented as a "bag of n-grams". Different lengths of n-grams such as a combination of 2-gram, 3-gram, 4-gram, and 5-grams have been experimented with. The experimental results show that the combinations of 2-gram and 4-gram features are the best character level n-gram features for the binary classification task. 

For the English datasets, they combined all the training datasets (\textsc{News}, \textsc{Wikinews}, and \textsc{Wikipedia}), used 3-gram, 4-gram and 5-gram character level n-gram features in order to maximize performance. The results show that character level n-gram features do not work well for cross-language complex word identification as the performance generally degraded.

For English, two variants of results were submitted, one classified using the corresponding in-domain training corpus and the second one classified using the concatenated training data. For German and Spanish, one result was submitted using the corresponding training data sets. For French, four submissions were made 1) one classified with English Wikipedia training, 2) one classified with all three English datasets, 3) one classified with Spanish data, and 4) one classified with German data. 

\textbf{NILC} present systems \cite{syspaper4} for the monolingual binary and probabilistic classification tasks. Three approaches were created by 1) using traditional feature engineering-based machine learning methods, 2) using the average embedding of target words as an input to a neural network, and 3) modeling the context of the target words using an LSTM. 

For the feature engineering-based systems, features such as linguistic, psycholinguistic, and language model features were used to train different binary and probabilistic classifiers. Lexical features include word length, number of syllables, and number of senses, hypernyms, and hyponyms in WordNet. For N-gram features, probabilities of the n-gram containing the target words were computed based on language models trained on the BookCorpus dataset and One Billion Word dataset. Furthermore, psycholinguistic features such as familiarity, age of acquisition, correctness and imagery values were used. Based on these features (38 in total), models were trained using Linear Regression, Logistic Regression, Decision Trees, Gradient Boosting, Extra Trees, AdaBoost, and XGBoost classifiers.

For embedding-based systems, a pre-trained GloVe model \cite{pennington-socher-manning:2014:EMNLP2014} was used to get the vector representations of target words. For MWE, the average of the vectors is used. In the first approach, the resulting vector is passed on to a neural network with two ReLu layers followed by a sigmoid layer, which predicted the probability of the target word being complex. 


Their experiments show that the feature engineering approach achieved the best results using the XGBoost classifier for the binary classification task. They submitted four systems using XGBoost, average embeddings, LSTMs with transfer learning, and a voting system that combines the other three. For the probabilistic classification task, their LSTMs achieve the best results.

\textbf{TMU} submitted multilingual and cross-lingual CWI systems for both of the binary and probabilistic classification tasks \cite{syspaper5}. The systems use two variants of frequency features from the learner corpus (\emph{Lang-8 corpus}) from \citet{mizumoto-EtAl:2011:IJCNLP-2011} and from the general domain corpus (Wikipedia and WikiNews). The list of features used in building the model include the number of characters in the target word, number of words in the target phrase, and frequency of the target word in learner corpus (Lang-8 corpus) and general domain corpus (Wikipedia and WikiNews). 

Random forest classifiers are used for the binary classification task while random forest regressors are used for the probabilistic classification task using the scikit-learn library. Feature ablation shows that both the length, frequency, and probability features (based on corpus statistics) are important for the binary and probabilistic classification tasks. They also discover that features obtained from the learner corpus are more influential than the general domain features for the CWI tasks. The systems perform very well both for the binary and probabilistic classification tasks, winning 5 out of the 12 tracks.

\textbf{ITEC} addresses both the binary and probabilistic classification task for the English and Spanish multilingual datasets \cite{syspaper6}. They have used 5 different aspects of the target word in the process of feature extractions, namely, word embedding, morphological structure, psychological measures, corpus counts, and topical information. Psychological measures are obtained from the MRC Psycholinguistic Database, which includes age of acquisition, imageability, concreteness, and meaningfulness of the target word. Word frequencies and embedding features are computed based on a web corpus. The word embedding model is computed using the gensim implementation of word2vec, with 300 dimensional embedding space, window-size of 5 and minimum frequency threshold of 20. 

They have employed deep learning structure using the keras deep learning library with the tensorflow gpu as a backend. Word embeddings are employed in two input layers, first to replace target words with the appropriate embeddings and second to represent the entire sentences as an input sequence which is considered the topical approximation using contextual cues. The final layer takes into account morphological features based on character embeddings that are trained with a convolutional network. The systems perform reasonably better than the average systems, for both of the binary and probabilistic classification tasks.

\begin{table*}[ht!]
\centering
\scalebox{0.90}{
   \begin{tabular}{|l|c|c|l|c|c|l|c|c|}
   \hline
   \hline
News & MAE & Rank & WikiNews & MAE & Rank & Wikipedia & MAE & Rank\\
\hline
		TMU & 0.051 & 1 & Camb & 0.0674 & 1 & Camb & 0.0739 & 1\\
		ITEC & 0.0539 & 2 & Camb & 0.0674 & 1 & Camb & 0.0779 & 2\\
		Camb & 0.0558 & 3 & Camb & 0.0690 & 2 & Camb & 0.0780 & 3\\
		Camb & 0.056 & 4 & Camb & 0.0693 & 3 & Camb & 0.0791 & 4\\
		Camb & 0.0563 & 5 & TMU & 0.0704 & 4 & ITEC & 0.0809 & 5\\
		Camb & 0.0565 & 6 & ITEC & 0.0707 & 5 & NILC & 0.0819 & 6\\
		NILC & 0.0588 & 7 & NILC & 0.0733 & 6 & NILC & 0.0822 & 7\\
		NILC & 0.0590 & 8 & NILC & 0.0742 & 7 & Camb & 0.0844 & 8\\
		SB@GU & 0.1526 & 9 & Camb & 0.0820 & 8 & TMU & 0.0931 & 9\\
		Gillin Inc. & 0.2812 & 10 & SB@GU & 0.1651 & 9 & SB@GU & 0.1755 & 10\\
		Gillin Inc. & 0.2872 & 11 & Gillin Inc. & 0.2890 & 10 & NILC & 0.2461 & 11\\
		Gillin Inc. & 0.2886 & 12 & Gillin Inc. & 0.3026 & 11 & Gillin Inc. & 0.3156 & 12\\
		NILC & 0.2958 & 13 & Gillin Inc. & 0.3040 & 12 & Gillin Inc. & 0.3208 & 13\\
		NILC & 0.2978 & 14 & Gillin Inc. & 0.3044 & 13 & Gillin Inc. & 0.3211 & 14\\
		Gillin Inc. & 0.3090 & 15 & Gillin Inc. & 0.3190 & 14 & Gillin Inc. & 0.3436 & 15\\
		SB@GU & 0.3656 & 16 & NILC & 0.3203 & 15 & NILC & 0.3578 & 16\\
		NILC & 0.6652 & 17 & NILC & 0.3240 & 16 & NILC & 0.3819 & 17\\
        \hline
		Baseline & 0.1127 & - & Baseline & 0.1053 & - & Baseline & 0.1112 & -\\
       \hline
   \end{tabular}
   }
 \caption{Probablistic classification results for the monolingual English tracks.} 
 \label{tab:Probresultsmono}
\end{table*}
     
 \begin{table*}[ht!]
\centering
\scalebox{0.90}{
   \begin{tabular}{|l|c|c|l|c|c|l|c|c|}
   \hline
   \hline
German & MAE & Rank & Spanish & MAE & Rank & French & MAE & Rank \\
\hline
TMU & 0.0610 & 1 & TMU & 0.0718 & 1 & CoastalCPH & 0.0660 & 1\\
CoastalCPH & 0.0747 & 2 & ITEC & 0.0733 & 2 & CoastalCPH & 0.0660 & 1\\
CoastalCPH & 0.0751 & 3 & CoastalCPH & 0.0789 & 3 & CoastalCPH & 0.0762 & 2\\
Gillin Inc. & 0.1905 & 4 & CoastalCPH & 0.0808 & 4 & TMU & 0.0778 & 3\\
Gillin Inc. & 0.2099 & 5 & Gillin Inc. & 0.2513 & 5 & CoastalCPH & 0.0866 & 4\\
Gillin Inc. & 0.2102 & 6 & Gillin Inc. & 0.2634 & 6 & - & - & - \\
Gillin Inc. & 0.2122 & 7 & Gillin Inc. & 0.2638 & 7 & - & - & - \\
- & - & - & Gillin Inc. & 0.2644 & 8 & - & - & - \\
- & - & - & CoastalCPH & 0.2724 & 9 & - & - & - \\
- & - & - & CoastalCPH & 0.2899 & 10 & - & - & - \\
\hline
Baseline & 0.0816 & - & Baseline & 0.0892 & - & Baseline & 0.0891 & -\\
    \hline
   \end{tabular}
   }
 \caption{Probablistic classification results for the multilingual German, Spanish, and French tracks.} 
 \label{tab:Probresultsmult}
\end{table*}

\textbf{Camb} describes different systems \cite{syspaper7} they have developed for the monolingual English datasets both for the binary and probabilistic classification tasks. They have used features that are based on the insights of the CWI shared task 2016 \cite{paetzold-specia:2016:SemEval1} such as lexical features (word length, number of syllables, WordNet features such as the number of synsets), word n-gram and POS tags, and dependency parse relations. In addition, they have used features such as the number of words grammatically related to the target word, psycholinguistic features from the MRC database, CEFR (Common European Framework of Reference for Languages) levels extracted from the Cambridge Advanced Learner Dictionary (CALD), and Google N-gram word frequencies using the Datamuse API The MCR features include word familiarity rating, number of phonemes, thorndike-lorge written frequency, imageability rating, concreteness rating, number of categories, samples, and written frequencies, and age of acquisition.

For the binary classification task, they have used a feature union pipeline to combine the range of heterogeneous features extracted from different categories of feature types. The best performing classification algorithms are obtained based on the ensemble techniques where AdaBoost classifier with 5000 estimators achieves the highest results, followed by the bootstrap aggregation classifier of Random Forest. All the features are used for the \textsc{News} and \textsc{WikiNews} datasets, but for the \textsc{Wikipedia} dataset, MCR psycholinguistic features are excluded. For the probabilistic classification task, the same feature setups are used and the Linear Regression algorithm is used to estimate values of targets. 

As it can be seen from Tables \ref{tab:binresultsmono}, \ref{tab:binresultsmulti}, \ref{tab:Probresultsmono}, and \ref{tab:Probresultsmult}, most of the systems submitted ranked first for English monolingual binary and probabilistic classification tasks.

\textbf{CoastalCPH} describe systems developed for multilingual and cross-lingual domains for the binary and probabilistic classification tasks \cite{syspaper8}. Unlike most systems, they have focused mainly on German, Spanish, and French datasets in order to investigate if multitask learning can be applied to the cross-lingual CWI task. They have devised two models, using language-agnostic approach with an ensemble that comprises of Random Forests (random forest classifiers for the binary classification task and random forest regressors for the probabilistic classification tasks, with 100 trees) and feed-forward neural networks. 

Most of the features are similar for all languages except some of them are language-specific features. The set of features incorporated include 1) log-probability features: unigram frequencies as a log-probabilities from language-specific Wikipedia dumps computed using KenLM, character perplexity, number of synsets, hypernym chain. 2) Inflectional complexity: number of suffixes appended to a word stem. 3) Surface features: length of the target and lower-case information. 4) Bag-of-POS: for each tag based on Universal Parts-of-Speech project, count the number of words in a candidate that belong to the respective class. 5) Target-sentence similarity: the cosine similarity between averaged word embeddings for the target word or phrase and the rest of the words in the sentence where out-of-vocabulary problems are addressed using a pre-trained sub-word embeddings \cite{DBLP:journals/corr/abs-1710-02187}.

 They have made qualitative and quantitative error analysis, mainly for the cross-lingual French dataset experiments and reported that: 1) The system picks longer targets as positive examples. 2) Short targets are predicted as false negative but they are potentially unknown named entities and technical terms. 3) Complex words are generally longer than simple words. 4) Language models produce lower log-probability for complex words. 
 
 The systems submitted performed the best out of all systems for the cross-lingual task (the French dataset) both for the binary and probabilistic classification tasks, showing a promising direction in the creation of CWI dataset for new languages.
 
\textbf{LaSTUS/TALN} present systems for the monolingual English binary classification task \cite{syspaper9}. Two different systems are designed, the first system is based on a set of lexical, semantic and contextual features, and the second system incorporates word embedding features. The word embedding features are obtained from a pre-trained word2vec model\footnote{https://code.google.com/archive/p/word2vec/}. 

For each sentence, the centroid of the dimensions of the context before the target word, the target word itself, and the context after the target word are computed using word2vec embedding vectors (300 dimensions each), resulting in a total of 900 feature dimensions. Furthermore, two extra features are generated using the embedding vectors, which represent the distance between the target word and the context before and after the target word respectively. These features are computed using the cosine similarity measures between each pair of the vectors.

A large set of shallow lexical and semantic features are also used in addition to the embedding features. These features include target word length (number of characters), the position of the target word in the sentence, number of words in the sentence, word depth in the dependency tree, parent word length in dependency relation, frequency features based on the BNC, Wikipedia, and Dale and Chall list corpora, number of synsets and senses in WordNet, and so on. 

The experiment is conducted using the Weka machine learning framework using the Support vector machine (with linear and radial basis function kernels), Na{\"i}ve Bayes, Logistic Regression, Random Tree, and Random Forest classification algorithms. The final experiments employ Support Vector Machines and Random Forest classifiers. 

\textbf{CFILT\_IITB} Developed ensemble-based classification systems for the English monolingual binary classification task \cite{syspaper10}. Lexical features based on WordNet for the target word are extracted as follows: 1) Degree of Polysemy: number of senses of the target word in WordNet, 2) Hyponym and Hypernym Tree Depth: the position of the word in WordNet's hierarchical tree, and 3) Holonym and Meronym Counts: based on the relationship of the target word to its components (meronyms) or to the things it is contained in (Holonym's). Additional feature classes include size-based features such as word count, word length, vowel counts, and syllable counts. They also use vocabulary-based features such as Ogden Basic (from Ogden's Basic Word list), Ogden Frequency (Ogden's Frequent Word List), and Barron's Wordlist (Barron's 5000 GRE Word List). 

They have used 8 classifiers namely Random Forest, Random Tree, REP Tree, Logistic Model Tree, J48 Decision Tree, JRip Rules Tree, PART, and SVM. Using these classifiers, a hard voting approach is used to predict a label for the target word. Voting of the positive or negative class is decided if more than 4 classifiers agree on the label. Word-embedding-based classifier is used to decide in the case of a 4-4 tie. 

An ablation test shows that size-based features such as word length, vowel counts, and syllable counts, word counts constitute the four top important features. Their best system shows an average performance compared to the other systems in the shared task for the monolingual English binary classification track.
 
\textbf{NLP-CIC} present systems for the English and Spanish multilingual binary classification tasks \cite{syspaper11}. The feature sets include morphological features such as frequency counts of target word on large corpora such as Wikipedia and Simple Wikipedia, syntactic and lexical features, psycholinguistic features from the MRC psycholinguistic database and entity features using the OpenNLP and CoreNLP tools, and word embedding distance as a feature which is computed between the target word and the sentence. 

Tree learners such as Random Forest, Gradient Boosted, and Tree Ensembles are used to train different classifiers. Furthermore, a deep learning approach based on 2D convolutional (CNN) and word embedding representations of the target text and its context is employed. 

Their best system ranked 10\textsuperscript{th}, 5\textsuperscript{th}, and 16\textsuperscript{th} for the \textsc{News}, \textsc{WikiNews}, and \textsc{Wikipedia} monolingual English tracks, which is better than the average systems in the shared task. The system based on the CNN model on the Spanish monolingual dataset ranked 2\textsuperscript{nd}.
	
\section{Conclusions}

This paper presented the results and findings of the second CWI shared task. Thirty teams enrolled to participate in the competition and 12 of them submitted their results. Subsequently, 11 teams wrote system description papers that have been reviewed in this report. 

Overall, traditional feature engineering-based approaches (mostly based on length and frequency features) perform better than neural network and word embedding-based approaches. However, compared to the SemEval 2016 Task 11 shared task systems presented in Table \ref{tab:approaches}, we have observed that more systems employed deep learning approaches and the results are getting better for the CWI task; the difference is less pronounced for the probabilistic classification tasks.

One of our most important findings is that cross-lingual experimental results are very promising, which we think implies in fundamental progress for CWI research. Despite the fact that we do not provide a training dataset for French, the results obtained have superior or equivalent scores (though they of course cannot be directly compared) to the German and Spanish datasets, when the system uses either one or several training datasets from the other languages.




\section*{Acknowledgments}

We would like to thank all participants of the CWI shared task, as well as the BEA workshop organizers for hosting and providing all the necessary support for the organization of the shared task. The dataset collection was funded as part of the DFG-SemSch project (BI-1544/3). 


\bibliographystyle{acl_natbib}
\bibliography{CWI}

\end{document}